\documentclass{article}

\usepackage{microtype}
\usepackage{graphicx}
\usepackage{subfigure}
\usepackage{booktabs} 

\usepackage{algorithm,algorithmic}
\usepackage{natbib}

\usepackage[accepted,nohyperref]{icml2019}
\newtheorem{theorem}{Theorem}

\ifx\assumption\undefined
\newtheorem{assumption}{Assumption}
\fi

\ifx\proposition\undefined
\newtheorem{proposition}[theorem]{Proposition}
\fi

\ifx\remark\undefined
\newtheorem{remark}{Remark}
\fi

\ifx\definition\undefined
\newtheorem{definition}{Definition}
\fi

\ifx\proof\undefined
\newenvironment{proof}{\par\noindent{\bf Proof\ }}{\hfill\BlackBox\\[2mm]}
\fi

\newcommand{\RN}[1]{%
	\textup{\lowercase\expandafter{\it \romannumeral#1}}%
}

\input{Definitions}
\icmltitlerunning{On Norm-Agnostic Robustness of Adversarial Training}

\begin{document}

\twocolumn[
\icmltitle{On Norm-Agnostic Robustness of Adversarial Training}




\begin{icmlauthorlist}
\icmlauthor{Bai Li}{sduke}
\icmlauthor{Changyou Chen}{b}
\icmlauthor{Wenlin Wang}{eduke}
\icmlauthor{Lawrence Carin}{eduke}
\end{icmlauthorlist}

\icmlaffiliation{sduke}{Department of Statistical Science, Duke University, Durham}
\icmlaffiliation{b}{Department of CSE, University at Buffalo, the State University of New York}
\icmlaffiliation{eduke}{Department of ECE, Duke University, Durham}

\icmlcorrespondingauthor{Bai Li}{bai.li@duke.edu}
\icmlcorrespondingauthor{Lawrence Carin}{lcarin@duke.edu}

\icmlkeywords{Machine Learning, ICML}

\vskip 0.3in
]



\printAffiliationsAndNotice{\icmlEqualContribution} 

\begin{abstract}
    Adversarial examples are carefully perturbed inputs for fooling machine learning models. A well-acknowledged defense method against such examples is adversarial training, where adversarial examples are injected into training data to increase robustness. In this paper, we propose a new attack to unveal an undesired property of the state-of-the-art adversarial training, that is it fails to obtain robustness against perturbations in $\ell_2$ and $\ell_\infty$ norms simultaneously. We discuss a possible solution to this issue and its limitations as well.
\end{abstract}

\section{Introduction}
  Deep neural networks (DNNs) have achieved significant success when applied to a variety of challenging machine-learning tasks. For example, DNNs have obtained state-of-the-art accuracy on large-scale image classification \citep{he2016identity,huang2017densely}. At the same time, vulnerability to adversarial examples, an undesired property of DNNs, has drawn attention in the deep-learning community \citep{szegedy2013intriguing,goodfellow2014explaining}. Generally speaking, adversarial examples are perturbed versions of the original data that successfully fool a classifier. For example, in the image domain, adversarial examples are images transformed from natural images with visually negligible changes but that lead to wrong predictions \citep{goodfellow2014explaining}. The existence of adversarial examples has raised many concerns, especially in scenarios with a high risk of misclassification, such as autonomous driving.
  
  To tackle adversarial examples, a variety of defensive methods against adversarial attacks have been proposed, yet most of them remain vulnerable to adaptive attacks \cite{szegedy2013intriguing,goodfellow2014explaining,moosavi2016deepfool,papernot2016distillation,kurakin2016adversarial,carlini2017towards,brendel2017decision,athalye2018obfuscated}. One type of adversarial defense that demonstrated good performance against strong attacks is based on adversarial training \citep{goodfellow2014explaining,madry2017towards, zhang2019theoretically}. Adversarial training constructs a defense model by augmenting the training set with adversarial examples. Though this is a simple strategy, it has achieved a great success in adversarial defense.
  
  The strength of attacks are commonly quantified by the $\ell_p$ distance between adversarial examples and natural examples. One desired property of a defense model is norm-agnostic, which requires a model to be robust against attacks constrained by a variety of norms. Recently, a more general attack mechanism called unrestricted adversarial attacks are introduced by \citet{brown2018unrestricted}, where adversarial examples are not necessarily close to a natural image as long as they are semantically similar. To achieve robustness against unrestricted attacks, being norm-agnostic is a minimum requirement.
  
  In this paper, we propose a new attack method and show adversarial training, the most successful adversarial defense models, is not norm-agnostic. Previously, it was reported both in \cite{madry2017towards} and \cite{zhang2019theoretically} that $\ell_\infty$ adversarial training is robust against $\ell_2$ attacks. Our experiments, however, suggest they fail to defend against $\ell_2$ and $\ell_\infty$ adversarial examples simultaneously.
  
  \section{Background and Related Work}
  
   Adversarial training constructs adversarial examples that are included to the training set to train a new and more robust classifier. This method is intuitive and has gained great success in defense \citep{szegedy2013intriguing,goodfellow2014explaining,madry2017towards,zhang2019theoretically}. \citet{madry2017towards} showed that iterative attacks during training yield strong defense models to while-box attacks \cite{athalye2018obfuscated}. More recently, another adversarial training based defense model \cite{zhang2019theoretically} has won the first place in the defense track of the NIPS 2018 Adversarial Vision Challenge \cite{brendel2018adversarial}.
   
   Although adversarial training has been so far one of the most successful defense methods, it has its limitations. In \cite{tramer2017ensemble}, it was pointed out that single-step adversarial training, where single-step method (e.g., FGSM \cite{goodfellow2014explaining}) is used for constructing adversarial examples, suffers from the ``degenerate global minimum'' issue and thus is not robust. To mitigate this issue, they propose ensemble adversarial training to improve the generalization of adversarial training. More recently, \cite{song2018improving} suggests using domain adaption as an improvement of ensemble adversarial training, leading to better robustness. However, both works only focus on single-step attack based adversarial training, while the most advanced adversarial training models are based on multi-step attacks. \citet{tramer2017ensemble} states that incorporating multi-step attacks during training could fix the degenerate-global-minimum issue. In this paper, we show multi-step adversarial training still suffers from this issue.

\section{Preliminary}
\subsection{Adversarial Examples}
    Given a classifier $f:\mathcal{X}\rightarrow \{1,\dots,k\}$ for an image $\mathbf{x}\in \mathcal{X}$, an adversarial example $\mathbf{x}_{\text{adv}}$ satisfies $\mathcal{D}(\mathbf{x},\mathbf{x}_{\text{adv}})<\epsilon$ for some small $\epsilon>0$, and $f(\mathbf{x})\neq f(\mathbf{x}_{\text{adv}})$, where $\mathcal{D}(\cdot, \cdot)$ is some distance metric, {\it i.e.}, $\mathbf{x}_{\text{adv}}$ is close to $\mathbf{x}$ but yields a different classification result. The distance is often described in terms of an $\ell_p$ metric, and in most of the literature $\ell_2$ and $\ell_\infty$ metrics are considered.

    One of the simplest and widely used attack methods is a single-step method, the Fast Gradient Sign Method (FGSM) \citep{kurakin2016adversarial}, which manipulates inputs along the direction of the gradient with respect to the outputs:
    \begin{equation}\label{eq:fgsm}
    \mathbf{x}_{\text{adv}}=\Pi_{\mathbf{x}+\mathcal{S}}(\mathbf{x}+\alpha (\nabla_\mathbf{x}L(\theta,\mathbf{x},y))
    \end{equation}
    where $\Pi_{\mathbf{x}+\mathcal{S}}$ is the projection operation that ensures adversarial examples stay in the $\ell_p$ ball $\mathcal{S}$ around $\mathbf{x}$.
    
    Its multi-step variant FGSM$^k$ is more powerful and has been shown to be equivalent to exploring adversarial examples with the projected gradient descent (PGD) method \cite{madry2017towards}: 
    \begin{equation}\label{eq:pgd}
    \mathbf{x}^{t+1}_{\text{adv}}=\Pi_{\mathbf{x}+\mathcal{S}}(\mathbf{x}^t_{\text{adv}}+\alpha (\nabla_\mathbf{x}L(\theta,\mathbf{x},y))
    \end{equation}

\subsection{Adversarial Training}

    The motivation behind adversarial training is that finding a robust model against adversarial examples is equivalent to solving the saddle-point problem: $$\min_\theta\max_{\mathbf{x}^\prime: D(\mathbf{x},\mathbf{x}^\prime)<\epsilon} L(\theta,\mathbf{x}^\prime,y)$$
    The inner maximization is equivalent to constructing adversarial examples, while the outer minimization can be performed by standard training procedure for loss minimization.
    
    Therefore, to achieve robustness to adversarial examples, adversarial training augments the training data with adversarial examples constructed during training, as an approximation to the inner maximization procedure.
    
    Recently, \citet{zhang2019theoretically} suggested using 
    $L(\theta,\mathbf{x},y)=L(f_\theta(\mathbf{x}),y)+\lambda L(f_\theta(\mathbf{x}),f_\theta(\mathbf{x}_{\text{adv}}))$ as the training loss, instead of $L(f_\theta(\mathbf{x}),y)$ and $L(f_\theta(\mathbf{x}_\text{adv}),y)$ alternatively used in \cite{madry2017towards}. 
    
    \subsection{Degenerate Global Minimum}
    
    In \cite{tramer2017ensemble}, it is pointed out that if $\mathbf{x}_{\text{adv}}$ denotes the adversarial example generated by FGSM, adversarial training ideally results in a robust classification model $\theta^*$ such that:
    $$L(\theta^*,\mathbf{x}_{\text{adv}},y)\approx \max_{\mathbf{x}^\prime: D(\mathbf{x},\mathbf{x}^\prime)<\epsilon} L(\theta^*,\mathbf{x}^\prime,y)\approx 0$$

    However, the training procedure may instead discover a ``degenerate global minimum'' $\theta^*$:
    $$L(\theta^*,\mathbf{x}_{\text{adv}},y)\ll \max_{\mathbf{x}^\prime: D(\mathbf{x},\mathbf{x}^\prime)<\epsilon} L(\theta^*,\mathbf{x}^\prime,y)$$

    In another word, the training procedure may generate a model that makes finding adversarial examples difficult for FGSM instead of a truly robust model.
    
    \cite{tramer2017ensemble} proposes two possible solutions for mitigating this issue. One is to use a strong multi-step adversarial training, such as PGD, at a cost of increased computational burden. Another is ensemble adversarial training, that is incorporating adversarial examples generated from multiple pre-trained classifiers that are different from the original one. In this way, they can decouple the construction of adversarial examples and the training to prevent ``degenerate global minimum'', while still obtain the robustness of adversarial training due to the transferability of adversarial perturbations across models \cite{goodfellow2014explaining}.

    \section{Second Order Attack}
     \label{sec:so}
    We propose a new attack motivated by the ``degenerate global minimum''. Note adversarial training is equivalent to solving the optimization problem: $$(\hat{\theta},\hat{\mathbf{x}})=\argmin_\theta \argmax_{\mathbf{x}^\prime:D(\mathbf{x},\mathbf{x}^\prime)<\epsilon} L(\theta,\mathbf{x}^\prime,y).$$ Its solution is a saddle point of $L$, {\it i.e.}, the gradient ideally vanishes at $\hat{\mathbf{x}}$ as $\nabla_\mathbf{x} L(\hat{\theta},\mathbf{x},y)|_{\hat{\mathbf{x}}}=0$. In practice, an adversarial training often finds $\hat{\theta}$ that makes the loss function flat in the neighborhood of a natural example $\mathbf{x}$, which leads to inefficient exploration for adversarial examples when performing attacks. This is intuitively the cause of ``degenerate global minimum''.
    
    \begin{table*}[ht]
        \vspace{-.4cm}
        \caption{Accuracy of Various Adversarial Training Strategies against Various Attacks}
        \label{T:adv_train}
        \begin{center}
        \begin{tabular}{ c || p{7.5mm}  p{7.5mm}  p{7.5mm}| p{7.5mm}  p{7.5mm}  p{7.5mm}|p{7.5mm}  p{7.5mm}  p{7.5mm}|p{7.5mm}  p{7.5mm}  p{7.5mm}}
        \hline
        &\multicolumn{3}{ c }{\textbf{Madry's}}  & \multicolumn{3}{c}{\textbf{TRADES }} &\multicolumn{3}{c}{\textbf{Ensemble }}&\multicolumn{3}{c}{\textbf{ATDA }}\\ 
        \cline{1-13}
        Attacks &$\ell_2$ & $\ell_\infty$ & Mix & $\ell_2$ & $\ell_\infty$ & Mix &$\ell_2$ & $\ell_\infty$ & Mix & $\ell_2$ & $\ell_\infty$ & Mix\\\hline
         Natural & 98.2\% & 98.8\% & 98.7\% & 99.4\% & 99.5\% & 99.4\% & 99.4\% & 99.0\% &98.7\% & 99.2\%&98.8\% & 99.0\%\\
         PGD ($\ell_2$) & 97.0\% & 92.8\% & 73.2\% & 91.7\% & 91.7\%& 90.4\% & 99.0\% & \textbf{65.3\%} & 58.2\% & 98.8\%& \textbf{63.6\%}&57.9\%\\
         PGD ($\ell_\infty$) & \textbf{0.4\%} & 92.5\% & 82.2\% & $\textbf{19.7\%}$ & 95.6\% & 15.3\% & \textbf{0.0}\% & 90.2\% &81.4\% &\textbf{0.0\%} &62.6\% & 81.3\%\\
         S-O ($\ell_2$)  & 96.6\% & \textbf{0.0\%} & 18.3\% & 81.7\%& \textbf{3.2\%} & 84.2\% & 98.9\% &\textbf{65.8\%} & 58.4\%& 97.2\%& \textbf{64.0\%} &56.8\%\\
         S-O ($\ell_\infty$) & \textbf{0.0\%} & 91.3\% &  84.2\% & \textbf{16.9\%} &94.7\% & 14.5\% & \textbf{0.0\%} & 88.9\%&82.1\%& \textbf{0.0\%}&61.3\%&83.4\%\\\hline
        \end{tabular}
        \vspace{-.4cm}
        \end{center}
    \end{table*}
    
    Most current attack methods construct adversarial examples based on the gradient of a loss function. However, according to the analysis above, first-order derivative is not effective for attacks if the defense model is trained adversarially. This motivates utilization of the second-order derivative of a loss function to construct adversarial examples.
   
    To this end, assume the loss function is twice differentiable with respect to $\mathbf{x}$. Using Taylor expansion on the difference between the losses on the original and perturbed samples, and assuming the gradient vanishes, we have
        $$L(\theta,\mathbf{x}+\mathbf{r},y) - L(\theta,\mathbf{x},y)\approx \frac{1}{2}\mathbf{r}^TH(\theta,\mathbf{x},y)\mathbf{r}$$
    with $\mathbf{r}$ being the perturbation, and $H(\theta,\mathbf{x},y)$ is the Hessian matrix of the loss function.
    Our goal is to find a small perturbation $\mathbf{r}$ that maximizes the difference $L(\theta,\mathbf{x}+\mathbf{r},y) - L(\theta,\mathbf{x},y)$. Our idea is based on the observation that the optimal perturbation direction should be in the same direction as the first dominant eigenvector, $\mathbf{e}(\theta,\mathbf{x},y)$, of $H(\theta,\mathbf{x},y)$, that is $\mathbf{r}=\epsilon \frac{\mathbf{e}(\theta,\mathbf{x},y)}{\|\mathbf{e}(\theta,\mathbf{x},y)\|_2}$ for some constant $\epsilon>0$. 
    However, computing the eigenvectors of the Hessian matrix requires $O(I^3)$ runtime with $I$ the dimension of the data. To tackle this issue, we adopt the fast approximation method from \cite{miyato2017virtual}, which is essentially a combination of the power-iteration method and the finite-difference method, to efficiently find the direction of the eigenvector.
    Based on this method, the optimal direction, denoted $\mathbf{r}_{adv}$, is approximated\footnote{Detailed derivations are provided in the Supplementary Material.} by
    \begin{equation}
        \label{eq:vat}
        \mathbf{r}_{adv} = \frac{g}{\|g\|_2},~~~~\text{with}~g=\nabla_{\mathbf{x}}L(\theta,\mathbf{x},y)|_{\mathbf{x}+\xi\mathbf{d}}
    \end{equation}
    where $\mathbf{d}$ is a randomly sampled unit vector and $\xi>0$ is a manually chosen step size. In practice, $\mathbf{d}$ is drawn from a centered Gaussian distribution and normalized such that its $\ell_2$ norm is $1$.
    
    This procedure is essentially a stochastic approximation to the optimal second-order direction, where the randomness comes from $\mathbf{d}$. To reduce
    the variance of the approximation, we further take the expectation over the Gaussian noise, yielding $g=\mathbb{E}_{\mathbf{d}\sim N(0,\sigma^2I)}\left[\nabla_{\mathbf{x}}L(\theta,\mathbf{x},y)|_{\mathbf{x}+\mathbf{d}}\right]$. Note that choosing $\sigma$ is equivalent to choosing the step size $\xi$ in (\ref{eq:vat}).
    Finally, we construct adversarial examples by an iterative update via PGD:
    \begin{equation}
    \label{eq:so}
    \mathbf{x}^{t+1}=\Pi_{\mathbf{x}+\mathcal{S}}(\mathbf{x}^t+\alpha\mathbf{r}_{adv}) = \Pi_{\mathbf{x}+\mathcal{S}}(\mathbf{x}^t+\alpha \frac{g^t}{\|g^t\|_2})
    \end{equation}
    where $g^t=\mathbb{E}_{\mathbf{d}\sim N(0,\sigma^2I)}\left[\nabla_{\mathbf{x}}L(\theta,\mathbf{x},y)|_{\mathbf{x}+\mathbf{d}}\right]$. Intuitively, this method perturbs the example at each iteration and tries to move out of the local maximum in the sample space, due to the introduction of random Gaussian noise.

\section{Experiments}
  \label{sec:exp}
  We perform experiments on the MNIST data set to validate our claims on adversarial training.
  
  The architecture of our model follows the ones used in \citep{madry2017towards}. Specifically, the model contains two convolutional layers with $32$ and $64$ filters, each followed by $2\times 2$ max-pooling, and a fully connected layer of size $1024$. Image intensities are scaled to $[0,1]$, and the size of attacks are also rescaled accordingly. In all the experiments, we bound the $\ell_2$ norm less than $4.0$ while the $\ell_\infty$ norm less than $0.3$. 
  
  We evaluate PGD and proposed S-O attacks on four settings: adversarial training with PGD adversarial examples \cite{madry2017towards}, Tradeoff-inspired Adversarial Defense (TRADES) \cite{zhang2019theoretically}, ensemble adversarial training \cite{tramer2017ensemble}, adversarial training via domain adaption\cite{song2018improving}. 
  
  Specifically, we first consider constructing adversarial examples with $\ell_2$ and $\ell_\infty$ constraints during training respectively. Table \ref{T:adv_train} shows, as expected, that Madry's model and TRADES successfully defend attacks with the same norm constraints. However, in spite of the fact that $\ell_\infty$ adversarial training stays robust against $\ell_2$ PGD attack, S-O attack can effectively reduce the accuracy of both models when a different norm is used. This suggests that standard adversarial training is not norm-agnostic. 
  
  It is natural to wonder whether the issue will be fixed if two kinds of adversarial examples are both included during training. To this end, we conduct additional experiments with mixed adversarial examples, that is alternating between $\ell_2$ and $\ell_\infty$ bounded examples for adversarial training. Using the mixed strategy, the accuracy on both attacks are no longer reduced to almost zero, but the overall performance is still unsatisfying. We conclude that mixing adversarial examples barely helps improving norm-agnostic robustness.
    
  According to our analysis, the poor performance of adversarial training is due to ``degenerate global minimum'', therefore, we expect ensemble adversarial training could fix the problem, as suggested in \cite{tramer2017ensemble}. The results from Table \ref{T:adv_train} suggest ensemble adversarial training and domain adaption partially fixes the issue, although the accuracy against $\ell_2$ attacks is still far from ideal.
  
  In addition, we found two more interest phenomenons that can support our claims. Firstly, the relatively good performance of ensemble adversarial training implies that the vulnerability to adversarial perturbations with different norms is indeed caused by the ``degenerate global minimum'' issue, similar to the single-step adversarial training. Secondly, the performance of PGD and S-O attacks become similar for ensemble adversarial training model. This implies the effectiveness of S-O attacks compared to PGD attacks is due to exploitation of the ``degenerate global minimum'' issue. 
    
\begin{figure}[ht]
    \centering
    \includegraphics[width=0.4\textwidth]{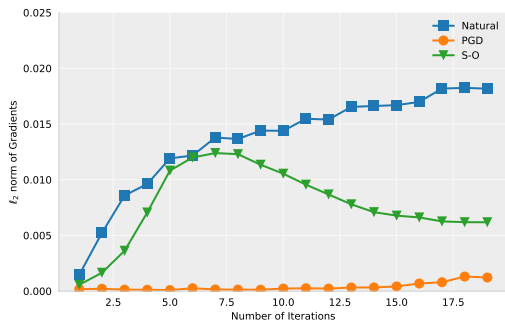}
    \caption{\textbf{S-O attack} Left: Average $\ell_2$ norm of the gradients of the loss function for a batch in each iteration during adversarial attack. \textbf{Blue}: Naturally trained model attacked by PGD. \textbf{Orange}: TRADES attacked by PGD. \textbf{Green}: TRADES attacked by S-O.}
    \label{fig:result1}
  \end{figure}

  In figure \ref{fig:result1}, we take a closer look at the behaviour of the attack methods by plotting the average $\ell_2$ norms of the gradients of the loss function with respect to the adversarial examples during the construction processes. Specifically, we compute $\frac{1}{N_{\text{batch}}}\sum_{i\in I}\|\nabla_{\mathbf{x}}L(\theta,\mathbf{x},y)|_{\mathbf{x}_i^{(t)}}\|_2$ for each $t$, where $I$ is the index set of a batch. 
  
  We monitor this quantity for three settings: naturally trained model attacked by PGD, TRADES attacked by PGD, and TRADES attacked by S-O. The difference between the blue and orange lines show the $\ell_2$ norms of the gradients of the adversarially trained model are much smaller than the ones of the naturally trained model under PGD attacks, validating our explanation in Section \ref{sec:so}, that an adversarially trained model tends to make the loss function ``flat'' in the neighborhood of natural examples which makes PGD attacks inefficient. The difference between the orange and green lines shows S-O attack is able to construct adversarial examples more efficiently by correctly finding the steepest direction, which explains why adversarially trained models are vulnerable to it.
    
\begin{figure}[ht]
        \centering
          \includegraphics[width=7cm]{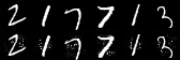}
        \caption{\textbf{Above:} Natural examples from MNIST. The correct labels are 2,1,7,7,1,3. \textbf{Below:} Adversarial examples with perturbation size $\ell_2\leq2.0$. The adversarially trained model predictions are 0,4,9,2,4,8.}
        \label{fig:illus}
      \end{figure}

    Finally, one may argue that the perturbation size $\ell_2=4.0$ is too large that it violates the assumption that adversarial perturbations are visually negligible. We therefore perform S-O attack with perturbation size $\ell_2\leq2.0$, which results in accuracy $41.04\%$. We also illustrate some randomly selected perturbed adversarial examples that are misclassified by TRADES in figure \ref{fig:illus}. 
    
    One can observe that although noticeable, there are only a limited amount of perturbations in the adversarial examples that do not change the semantic meaning of the images. 
    
     \paragraph{CIFAR-10} It is worth-noting that we do not observe similar results on CIFAR-10. We believe on CIFAR-10, it is difficult to reach even a ``degenerate global minimum'' for adversarial training due to the high dimensionality of the input space. This explains why adversarial training is still far from being perfectly robust even against PGD \cite{madry2017towards,zhang2019theoretically}.
    \section{Conclusion}
    
    In this paper, we show multi-step adversarial training models suffer from ``degenerate global minimum'' and thus are not norm-agnostic robust. Our proposed attack method is capable of constructing adversarial examples that reduces the accuracy of state-of-the-art adversarial training when different norms are used for training and attacking. 
    
    On the other hand, ensemble adversarial training can mitigate the issue thus should be considered as a standard procedure for adversarial training, even though they can only obtain moderate adversarial robustness.
    
    In general, considering state-of-the-art results in adversarial defense are often achieved by adversarial training, we believe it is important to check the norm-agnostic robustness when designing adversarial defense models. 
\bibliography{references.bbl}

\begin{thebibliography}{19}
\providecommand{\natexlab}[1]{#1}
\providecommand{\url}[1]{\texttt{#1}}
\expandafter\ifx\csname urlstyle\endcsname\relax
  \providecommand{\doi}[1]{doi: #1}\else
  \providecommand{\doi}{doi: \begingroup \urlstyle{rm}\Url}\fi

\bibitem[Athalye et~al.(2018)Athalye, Carlini, and
  Wagner]{athalye2018obfuscated}
Athalye, A., Carlini, N., and Wagner, D.
\newblock Obfuscated gradients give a false sense of security: Circumventing
  defenses to adversarial examples.
\newblock \emph{arXiv preprint arXiv:1802.00420}, 2018.

\bibitem[Brendel et~al.(2017)Brendel, Rauber, and Bethge]{brendel2017decision}
Brendel, W., Rauber, J., and Bethge, M.
\newblock Decision-based adversarial attacks: Reliable attacks against
  black-box machine learning models.
\newblock \emph{arXiv preprint arXiv:1712.04248}, 2017.

\bibitem[Brendel et~al.(2018)Brendel, Rauber, Kurakin, Papernot, Veliqi,
  Salath{\'e}, Mohanty, and Bethge]{brendel2018adversarial}
Brendel, W., Rauber, J., Kurakin, A., Papernot, N., Veliqi, B., Salath{\'e},
  M., Mohanty, S.~P., and Bethge, M.
\newblock Adversarial vision challenge.
\newblock \emph{arXiv preprint arXiv:1808.01976}, 2018.

\bibitem[Brown et~al.(2018)Brown, Carlini, Zhang, Olsson, Christiano, and
  Goodfellow]{brown2018unrestricted}
Brown, T.~B., Carlini, N., Zhang, C., Olsson, C., Christiano, P., and
  Goodfellow, I.
\newblock Unrestricted adversarial examples.
\newblock \emph{arXiv preprint arXiv:1809.08352}, 2018.

\bibitem[Carlini \& Wagner(2017)Carlini and Wagner]{carlini2017towards}
Carlini, N. and Wagner, D.
\newblock Towards evaluating the robustness of neural networks.
\newblock In \emph{Security and Privacy (SP), 2017 IEEE Symposium on}, pp.\
  39--57. IEEE, 2017.

\bibitem[Goodfellow et~al.(2014)Goodfellow, Shlens, and
  Szegedy]{goodfellow2014explaining}
Goodfellow, I.~J., Shlens, J., and Szegedy, C.
\newblock Explaining and harnessing adversarial examples.
\newblock \emph{arXiv preprint arXiv:1412.6572}, 2014.

\bibitem[He et~al.(2016{\natexlab{a}})He, Zhang, Ren, and Sun]{he2016deep}
He, K., Zhang, X., Ren, S., and Sun, J.
\newblock Deep residual learning for image recognition.
\newblock In \emph{Proceedings of the IEEE conference on computer vision and
  pattern recognition}, pp.\  770--778, 2016{\natexlab{a}}.

\bibitem[He et~al.(2016{\natexlab{b}})He, Zhang, Ren, and Sun]{he2016identity}
He, K., Zhang, X., Ren, S., and Sun, J.
\newblock Identity mappings in deep residual networks.
\newblock In \emph{European conference on computer vision}, pp.\  630--645.
  Springer, 2016{\natexlab{b}}.

\bibitem[Krizhevsky et~al.(2012)Krizhevsky, Sutskever, and
  Hinton]{krizhevsky2012imagenet}
Krizhevsky, A., Sutskever, I., and Hinton, G.~E.
\newblock Imagenet classification with deep convolutional neural networks.
\newblock In \emph{Advances in neural information processing systems}, pp.\
  1097--1105, 2012.

\bibitem[Kurakin et~al.(2016)Kurakin, Goodfellow, and
  Bengio]{kurakin2016adversarial}
Kurakin, A., Goodfellow, I., and Bengio, S.
\newblock Adversarial machine learning at scale.
\newblock \emph{arXiv preprint arXiv:1611.01236}, 2016.

\bibitem[Madry et~al.(2017)Madry, Makelov, Schmidt, Tsipras, and
  Vladu]{madry2017towards}
Madry, A., Makelov, A., Schmidt, L., Tsipras, D., and Vladu, A.
\newblock Towards deep learning models resistant to adversarial attacks.
\newblock \emph{arXiv preprint arXiv:1706.06083}, 2017.

\bibitem[Miyato et~al.(2017)Miyato, Maeda, Koyama, and
  Ishii]{miyato2017virtual}
Miyato, T., Maeda, S.-i., Koyama, M., and Ishii, S.
\newblock Virtual adversarial training: a regularization method for supervised
  and semi-supervised learning.
\newblock \emph{arXiv preprint arXiv:1704.03976}, 2017.

\bibitem[Moosavi-Dezfooli et~al.(2016)Moosavi-Dezfooli, Fawzi, and
  Frossard]{moosavi2016deepfool}
Moosavi-Dezfooli, S.-M., Fawzi, A., and Frossard, P.
\newblock Deepfool: a simple and accurate method to fool deep neural networks.
\newblock In \emph{Proceedings of the IEEE Conference on Computer Vision and
  Pattern Recognition}, pp.\  2574--2582, 2016.

\bibitem[Papernot et~al.(2016)Papernot, McDaniel, Wu, Jha, and
  Swami]{papernot2016distillation}
Papernot, N., McDaniel, P., Wu, X., Jha, S., and Swami, A.
\newblock Distillation as a defense to adversarial perturbations against deep
  neural networks.
\newblock In \emph{Security and Privacy (SP), 2016 IEEE Symposium on}, pp.\
  582--597. IEEE, 2016.

\bibitem[Song et~al.(2018)Song, He, Wang, and Hopcroft]{song2018improving}
Song, C., He, K., Wang, L., and Hopcroft, J.~E.
\newblock Improving the generalization of adversarial training with domain
  adaptation.
\newblock \emph{arXiv preprint arXiv:1810.00740}, 2018.

\bibitem[Szegedy et~al.(2013)Szegedy, Zaremba, Sutskever, Bruna, Erhan,
  Goodfellow, and Fergus]{szegedy2013intriguing}
Szegedy, C., Zaremba, W., Sutskever, I., Bruna, J., Erhan, D., Goodfellow, I.,
  and Fergus, R.
\newblock Intriguing properties of neural networks.
\newblock \emph{arXiv preprint arXiv:1312.6199}, 2013.

\bibitem[Tram{\`e}r et~al.(2017)Tram{\`e}r, Kurakin, Papernot, Goodfellow,
  Boneh, and McDaniel]{tramer2017ensemble}
Tram{\`e}r, F., Kurakin, A., Papernot, N., Goodfellow, I., Boneh, D., and
  McDaniel, P.
\newblock Ensemble adversarial training: Attacks and defenses.
\newblock \emph{arXiv preprint arXiv:1705.07204}, 2017.

\bibitem[Zagoruyko \& Komodakis(2016)Zagoruyko and
  Komodakis]{zagoruyko2016wide}
Zagoruyko, S. and Komodakis, N.
\newblock Wide residual networks.
\newblock \emph{arXiv preprint arXiv:1605.07146}, 2016.

\bibitem[Zhang et~al.(2019)Zhang, Yu, Jiao, Xing, Ghaoui, and
  Jordan]{zhang2019theoretically}
Zhang, H., Yu, Y., Jiao, J., Xing, E.~P., Ghaoui, L.~E., and Jordan, M.~I.
\newblock Theoretically principled trade-off between robustness and accuracy.
\newblock \emph{arXiv preprint arXiv:1901.08573}, 2019.

\end{thebibliography}
\bibliographystyle{plain}

\appendix

 \section{Fast Approximate Method \cite{miyato2017virtual}}
  Power iteration method \citep{golub2001eigenvalue} allows one to compute the dominant eigenvector $\mathbf{r}$ of a matrix $\mathbf{H}$. Let $\mathbf{d}^0$ be a randomly sampled unit vector which is not perpendicular to $\mathbf{r}$, the iterative calculation of
  
  $$\mathbf{d}^{t+1}=\frac{\mathbf{Hd}^{t}}{\|\mathbf{Hd}^{t}\|_2}$$
  
  leads to $\mathbf{d}^{t}\rightarrow \mathbf{r}$. Given $\mathbf{H}$ is the Hessian matrix of $L(\theta,\mathbf{x},y)$, we further use finite difference method to reduce the computational complexity:
  
  \begin{align*} 
    \mathbf{Hd}&\approx \frac{\nabla_{\mathbf{x}+\xi\mathbf{d}}L(\theta,\mathbf{x}+\xi\mathbf{d},y)-\nabla_{\mathbf{x}}L(\theta,\mathbf{x},y)}{\xi}\\
    &=\frac{\nabla_{\mathbf{x}+\xi\mathbf{d}}L(\theta,\mathbf{x}+\xi\mathbf{d},y)}{\xi}
  \end{align*}
  
  where $\xi>0$ is the step size. If we only take one iteration, it gives an approximation that only requires the first-order derivative:
  
  $$\mathbf{r}\approx \frac{\mathbf{Hd}}{\|\mathbf{Hd}^{t}\|_2}\approx \frac{\nabla_{\mathbf{x}+\xi\mathbf{d}}L(\theta,\mathbf{x}+\xi\mathbf{d},y)}{\|\nabla_{\mathbf{x}+\xi\mathbf{d}}L(\theta,\mathbf{x}+\xi\mathbf{d},y)\|}$$
  
  which gives equation \ref{eq:vat}.

\end{document}